\documentclass[10pt,twocolumn,letterpaper]{article}

\usepackage[pagenumbers]{cvpr} 

\usepackage{graphicx}
\usepackage{amsmath}
\usepackage{amssymb}
\usepackage{booktabs}
\usepackage{times}
\usepackage{epsfig}
\usepackage{graphicx}
\usepackage{amsmath}
\usepackage{amssymb}
\usepackage{times,enumitem}
\usepackage{epsfig}
\usepackage{graphicx}
\usepackage{amsmath}
\usepackage{amssymb}
\usepackage{mathtools}
\usepackage[table,xcdraw]{xcolor}
\usepackage{cuted}
\usepackage{capt-of}
\usepackage{comment}

\usepackage[pagebackref,breaklinks,colorlinks]{hyperref}

\usepackage[capitalize]{cleveref}
\crefname{section}{Sec.}{Secs.}
\Crefname{section}{Section}{Sections}
\Crefname{table}{Table}{Tables}
\crefname{table}{Tab.}{Tabs.}

\def\cvprPaperID{0987} 
\def\confName{CVPR}
\def\confYear{2022}

\begin{document}

\title{Fine-Context Shadow Detection using Shadow Removal}

\author{Jeya Maria Jose Valanarasu and  
Vishal M. Patel\\
Johns Hopkins University
}

\maketitle

\vspace{-3em}

\begin{abstract}

	Current shadow detection methods perform poorly when detecting shadow regions that are small, unclear or have blurry edges. In this work, we attempt to address this problem on two fronts. First, we propose a Fine Context-aware Shadow Detection Network (FCSD-
	Net), where we constraint the receptive field size and focus on low-level features to learn fine context features better. Second, we propose a new learning strategy, called Restore to Detect (R2D), where we show that when a deep neural network is trained for restoration (shadow removal), it learns meaningful features to delineate the shadow masks as well. To make use of this complementary nature of shadow detection and removal tasks, we train an auxiliary network for shadow removal and propose a complementary feature learning block (CFL) to learn and fuse meaningful features from shadow removal network to the shadow detection network. We train the proposed network, FCSD-Net, using the  R2D learning strategy across multiple datasets. Experimental results on three public shadow detection datasets (ISTD, SBU and UCF) show that our method improves the shadow detection performance while being able to detect fine context better compared to the other recent methods. 
	
\end{abstract}

\section{Introduction}
\label{intro}
	\begin{figure}
	\centering
	\includegraphics[width=0.45\textwidth]{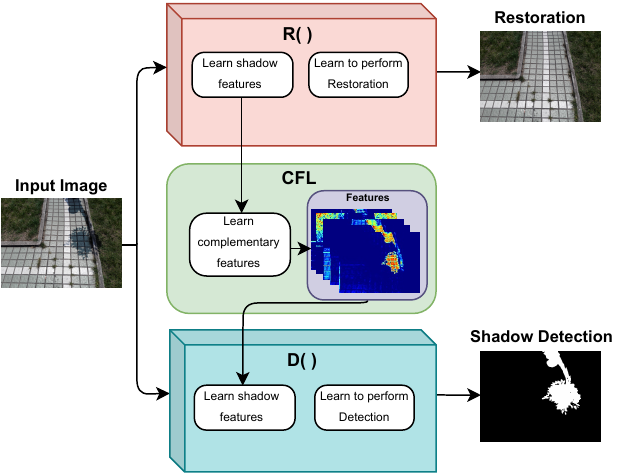}
	\caption{Our proposed learning strategy R2D. The shadow feature maps learned by the restoration network $R ()$ are forwarded to the detection network $D ()$ via a complementary feature learning block (CFL) to enhance the detection performance.}
	\label{intro}
	\vspace{-0.5em}
\end{figure}
There is a high probability of the presence of shadows in any image captured in natural conditions. Shadows are caused by objects occluding the light from the illumination source thus causing color, intensity and texture changes in the surface where the light is obstructed. Presence of a single object with an illumination source is enough to cast a shadow (excluding perpendicular conditions), be it outdoors or indoors. Detecting shadows is of high importance in computer vision because of two main reasons. First, shadows hamper the image scene causing a performance drop for other major vision tasks like semantic segmentation \cite{guan2008wavelet}, object detection \cite{mikic2000moving}, video surveillance \cite{matusek2008shadow} and visual tracking \cite{cucchiara2001improving}. In these tasks, careful delineation or removal of shadows aids in a performance boost. Second,  detecting shadows help in other tasks like inferring the scene geometry \cite{karsch2011rendering, junejo2008estimating, okabe2009attached}, camera parameters \cite{wu2010camera} and light source localization \cite{panagopoulos2009robust, lalonde2009estimating}.

Traditional methods proposed for shadow detection in images develop physical models \cite{finlayson2005removal, finlayson2009entropy} or use machine learning models based on hand-crafted features \cite{lalonde2010detecting, huang2011characterizes, zhu2010learning}. Following the popularity of deep learning methods in solving computer vision tasks, a lot of methods based on convolutional neural networks (ConvNets) were proposed for the task of shadow detection \cite{zhao2019egnet,zheng2019distraction,zhu2018bidirectional, wang2017stagewise,wang2018stacked,wang2019densely, chen2020multi, chen2018reverse, vicente2016large, vicente2017leave, nguyen2017shadow}. ConvNet-based methods are found to be successful for shadow detection as they are able to learn global contexts of the shadow region better than any other previous methods. This key property is specifically demonstrated in recent works \cite{zhu2018bidirectional,nguyen2017shadow, hu2018direction}. However, these methods perform poorly when shadow regions or small or have blurry boundaries. In this work, we focus on solving the fine-context problem in shadow detection.

 First, we propose a new network architecture for the detection  to efficiently segment shadow regions of fine context. Although ConvNets work better at capturing global context as the deeper layers focus on bigger objects due to a larger receptive field, the number of filters focusing on extracting information about fine context is still very less as only the shallow layers are responsible for extracting local information. Even though extracting local features were explored by using input at different resolutions \cite{zhu2018bidirectional}, it still does not work well in segmenting fine context shadow regions which are small or with unclear boundaries. To this end, we propose a Fine Context aware Shadow Detection Network (FCSD-Net) where we introduce a fine context detector block that constraints the receptive field size from enlarging in the deeper layers thus helping in feature extraction of shadow regions which are small, unclear or with confounding boundaries. 
 
\begin{figure}[t!]
	\centering
	\includegraphics[width=1\linewidth]{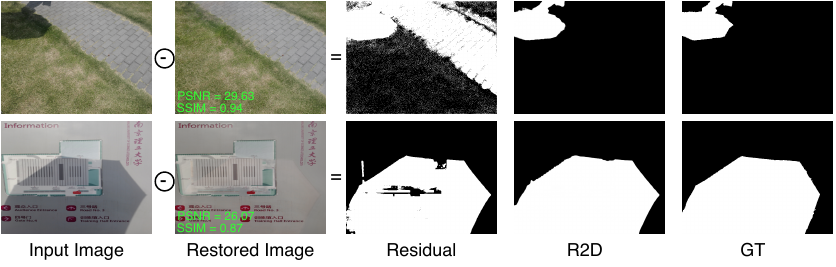}
	\vskip -5pt
	\caption{Predicting the shadow segmentation map by finding the residual of restored image from the input image. It can be seen that the residual is noisy while our proposed learning strategy R2D predicts a shadow segmentation as good as the ground truth. }
	\vspace{-1em}
	\label{shadresid}
	\vspace{-0.5em}
\end{figure}

Shadow removal is also a growing area of research in computer vision where the task is to restore the original image by removing the shadow. Traditional methods \cite{liu2008texture,xiao2007moving,baba2004shadow,arbel2010shadow} as well as deep learning-based methods \cite{le2019shadow,lin2020bedsr,vasluianu2020self,anand2019tackling,khan2015automatic,chen2021triplecooperative} have been proposed to solve this task. Shadow detection and removal can be thought of as complementary tasks as one involves detecting the shadow region and the other involves removing the shadow region and replacing it with the background scenery. A naive solution for shadow detection using shadow removal would be to subtract the restored image from the input image. However the residual is very noisy as the restoration is not perfect even if it looks visually good. An example for this can be found in Fig \ref{shadresid} where we restore the image using the current state-of-the-art shadow removal method--dual
hierarchically aggregation network (DHAN) \cite{cun2020towards}. It can be noted that even though the PSNR and SSIM are reasonable for restoration, the residual is still not the desirable shadow segmentation. Thus in this work, we propose a new way to efficiently use the removal task to enhance the detection task.

It is evident that deep networks trained for shadow removal also learn features of that about the shadow region. This concept in explained in detailed in section \ref{comp} and also illustrated in Fig \ref{featresponse-r2d}. We try to make use of this phenomenon and avail these feature maps to enhance shadow detection. A recently published dataset - ISTD \cite{wang2018stacked} consists of triplets of images- image with shadow, shadow segmentation and the clean image making this idea feasible. It can be noted that no previous method has attempted to make use of the restoration task to improve the detection task. The closest work is \cite{wang2018stacked} where a stacked conditional generative adversarial network was designed to solve shadow detection and removal in sequence. It adds the shadow detection mask as an input the removal network however its detection pipeline remains generic. We propose a new learning strategy for shadow detection, called  Restore to Detect (R2D), where we have a  restoration network $R ()$ that learns shadow removal as an auxiliary task and feeds the feature maps learned to the detection network $D ()$. We also propose a complementary feature leaning block (CFL) that learns to feed only the shadow features from $R ()$ to $D ()$. The proposed learning framework R2D is illustrated in Fig \ref{intro}. To the best of our knowledge, this is the first work that attempts to use the shadow removal task to enhance shadow detection. We perform extensive experiments on multiple datasets to show that our proposed network FCSD-Net with the proposed learning strategy R2D performs better then the existing methods both qualitatively and quantitatively.

The main contributions of this paper are as follows:

\begin{itemize}[noitemsep]

	\item We propose FCSD-Net that can effectively detect shadow regions which are unclear, blurry or small. It also reduces the false-detection of background regions with similar color and intensity as shadows.
	
	\item We propose a new learning strategy called Restore to Detect (R2D) that leverages the feature maps learned during shadow removal for enhancing the performance of shadow detection. This learning strategy can be easily adopted by any current or future shadow detection methods with ease. 
	
	\item We achieve improvements in performance across three public shadow detection datasets - SBU \cite{vicente2016large}, UCF \cite{zhu2010learning} and ISTD \cite{wang2018stacked} both quantitatively and qualitatively, especially in confounding cases.
\end{itemize}

\section{Related Work}

\noindent {\bf{Traditional methods. }} Early methods for shadow detection focused on hand crafting meaningful features to discriminate the shadow from the background in the image. For example, \cite{lalonde2010detecting,huang2011characterizes,zhu2010learning} design hand-crafted features based on edge and pixel information of shadow regions for shadow detection in consumer photographs. Later, various classifiers were explored using the hand-crafted features. In particular, Huang \textit{et al.} \cite{huang2011characterizes} used edge features of shadow to train an SVM \cite{hearst1998support} for shadow detection. Guo \textit{et al.} \cite{guo2011single} used a graph-based classifier to segment shadows using illumination features. Vicente \textit{et al.} \cite{vicente2017leave} used MRF to boost the shadow detection performance by using pairwise region context information. Few early works \cite{finlayson2009entropy,finlayson2005removal} also explored building illumination and color based models to segment shadows. Most of these methods perform poorly in complex scenarios as hand-crafted features do not discriminate shadow region from the image background well.

\noindent {\bf{Deep learning-based methods. }} 
Following the success of deep learning in various computer vision tasks such as semantic segmentation, object detection, image restoration, and image-to-image translation, various ConvNet-based methods for shadow detection have been proposed in the literature \cite{khan2014automatic, vicente2016large, nguyen2017shadow, le2018a+, hu2018direction}. Khan \textit{et al.} \cite{khan2014automatic} first used a simple 7 layer ConvNet to automatically extract the feature descriptors for shadow detection. The feature learning part was done at the super-pixel level and along the shadow boundaries. Vicente \textit{et al.} \cite{vicente2016large} proposed a semantic patch level ConvNet to train efficiently on patches while also using the image level semantic information. Nguyen \textit{et al.} \cite{nguyen2017shadow} introduced scGAN where cGAN was tailor-made for the task of shadow detection. scGAN used an adversarial approach to model  high-level context and global features better.  He \textit{et al.} \cite{le2018a+} proposed another adversarial strategy where two separate networks - attenuation network and shadow detection network were trained together in an adversarial way where the attenuation network tries to fool the detection network's shadow prediction by modifying the input shadow image. This effectively acts as data augmentation to  training data as the shadow detection network is now trained on various new instances of the input data.

Hu \textit{et al.} \cite{hu2018direction} explored a direction-aware manner to analyze image context for shadow detection. A new module, direction-aware spatial context (DSC), was proposed which uses a spatial RNN to learn spatial context of shadows. Zhu \textit{et al.} \cite{zhu2018bidirectional} proposed a bidirectional feature pyramid network (BFPN) which combines contexts of two different ConvNets by using a recurrent attention residual (RAR) to refine the context features from deep to shallow layers. Hosseinzadeg \textit{et al.} \cite{hosseinzadeh2018fast} proposed obtaining a shadow prior first using multi-class SVM using statistical features and then use it along with the original image to train a patch-level ConvNet. Zheng \textit{et al.} \cite{zheng2019distraction} proposed a Distraction aware Shadow Detection (DSDNet) where a standalone Distraction-aware Shadow (DS) module was introduced to learn discriminative features for robust shadow detection. DS module learns to predict the false positive and false negative maps explicitly which helps in learning and integrating the visual distraction regions to get an efficient shadow detection. Wang \textit{et al.} \cite{wang2019densely} proposed a densely cascaded learning  method to fuse both global and local details efficiently. Recently, Chen \textit{et al.} \cite{chen2020multi} proposed a semi-supervised framework using a teacher-student network where unlabeled shadow data were used to further improve the performance of the network. 

\section{Proposed Method}
In this section, we explain our proposed architecture FCSD-Net and learning strategy R2D in detail.

\subsection{Fine Context Shadow Detection}

\begin{figure*}
	\centering
	\includegraphics[width=0.8\linewidth, height = 0.6\linewidth]{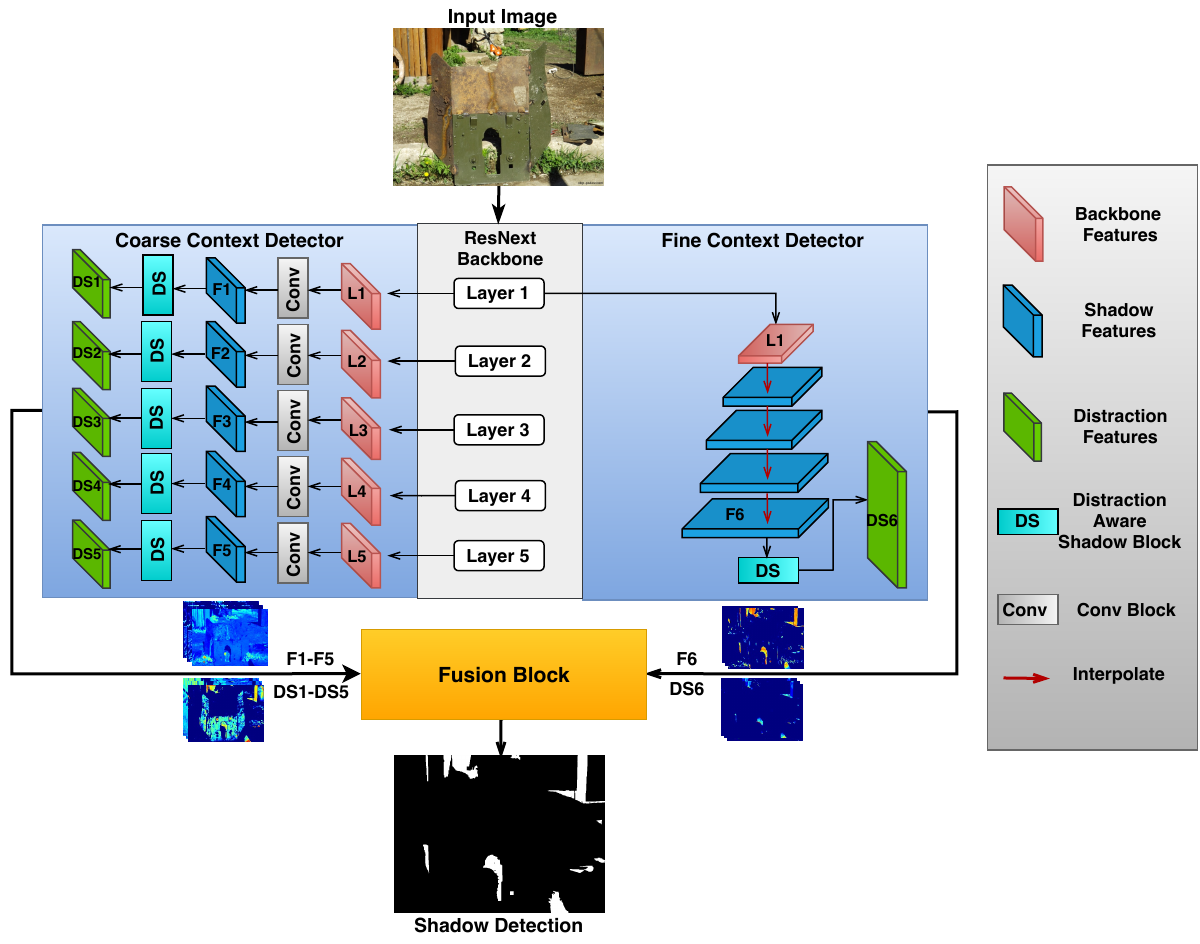}
	\caption{The proposed architecture for FCSD-Net. The input image is first fed to the ResNeXt backbone network. Backbone features $L1-L5$ are forwarded to the CCD and FCD blocks. The feature maps $F1-F6$ and $DS1-DS6$ extracted from these blocks are forwarded to the fusion block which outputs the shadow segmentation map. }
	\label{fcsd}
	\vspace{-0.5em}
\end{figure*}

In most of the previous deep learning-based solutions for shadow detection, the ConvNets used are of a generic encoder-decoder network architecture where the input image is taken to a lower dimension. The conv layers are designed such that the receptive field size of the successive layers increases through every conv layer. This forces the network to focus on high level information in deeper layers. Although this actually helps the networks learn shadow structures of people and large objects well, it reduces the focus the network gives to smaller shadow regions and shadows with unclear boundaries. This happens because smaller shadow regions and sharper edges need conv filters with smaller receptive field to learn features that extract them. So, in our proposed solution to this problem we make use of an alternate design of ConvNets to focus more on fine details.    

Consider a configuration of two conv layers in succession where $I$ be the input image, $F_1$ and $F_2$ be the feature maps extracted from the conv layers 1 and 2, respectively. As in a generic ConvNet, let there be max-pooling layer between these conv layers. In this case, the receptive field of conv layer 2 (to which $F_1$ is forwarded) on the input image would be $ 2 \times k \times 2 \times k$. However if we have an upsampling layer instead of the max-pooling layer, the receptive field would become  $ \frac{1}{2} \times k \times \frac{1}{2} \times k$ \cite{valanarasu2020kiu,valanarasu2020kiu1}. This helps in the alternative ConvNet architecture to learn more low-level information like edges and other finer details better.

\subsection{FCSD-Net}

\begin{figure}[t]
	\centering
	\includegraphics[width=1\linewidth]{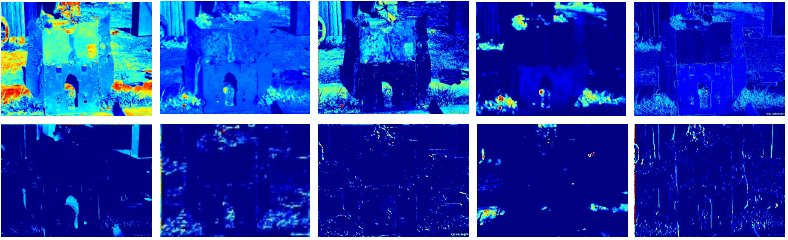}
	\caption{Feature maps collected from the trained FCSD-Net. Top row corresponds to the feature maps collected from CCD while the bottom row corresponds to the feature maps collected from FCD. It can be seen that feature maps in CCD focus on high-level context while the feature maps in FCD focus on fine context.}
	\label{featresponse-fcsd}
	\vspace{-2em}
\end{figure}

Using the above concept, we propose FCSD-Net which is an end-to-end trainable network where the input is the image with shadow and the output is shadow segmentation. Similar to \cite{zheng2019distraction,zhu2018bidirectional}, we use a ResNeXt backbone network for feature extraction. We select the output of the last conv layer of each conv block in the ResNeXT architecture. Specifically, we choose the features from the layers - conv1, res2c, res3b3, res4b22 and res5c and term them $L1, L2, L3, L4$ and $L5$ for simplicity. Note that, out of these feature maps $L1$ and $L2$ have low-level feature information while all the others have high-level information. From the backbone network, we pass the features extracted to two blocks - Fine context detector (FCD) and coarse context detector (CCD). We pass all the feature maps ($L1, L2, L3, L4 $ and $L5$) to CCD while we pass only $L1$ to FCD as the focus of FCD is to detect low-level shadow information.

In CCD, all these backbone feature maps are first forwarded to a conv block to learn the image features corresponding to shadow detection. We term these image features as $F1, F2, F3, F4 $ and $F5$, respectively. Then at each scale, we pass these image features to the Distraction Aware Shadow (DS) Block   \cite{zheng2019distraction}. We note from \cite{zheng2019distraction} that capturing the distraction features using the DS module helps in improving the shadow detection performance. It is designed such that the false positive distraction and the false negative distraction of the shadow are learned separately and then efficiently integrated. More information about the DS module can be found in \cite{zheng2019distraction}. We term these distraction aware features $DS1, DS2, DS3, DS4$ and $DS5$.    

In FCD, input $L1$ is forwarded to an alternative ConvNet architecture. Here, we have a set of 4 conv layers with upsampling layers in between them. Note that upsampling can be done either by interpolation techniques or transpose convolution. After some early experiments, we found that the performance of both methods were similar. So we use a simple bilinear interpolation in this work  to reduce the computation. The interpolation factor chosen at each layer is such that feature maps grow 50 more pixels in terms of height and width when compared to the feature map dimensions before the upsampling layer. The receptive field gets constrained after each upsampling layer and each successive conv layer. We term the fine context features extracted from the last conv layer as $F6$. Similar to CCD, we use the DS module here as well as it helps in capturing the shadow distraction features. We pass $F6$ to the DS module to get the distraction features $DS6$.

The feature maps $F1$ to $F6$, $DS1$ to $DS6$ taken from FCD and CCD are forwarded to the fusion block. In the fusion block, we interpolate all the feature maps to the same size as of the input image. After this we concatenate all the feature maps and pass them through a $1 \times 1$ conv layer. This output is forwarded to a sigmoid activation to get the binary shadow map as output. The architecture of FCSD-Net is illustrated in Fig~\ref{fcsd}. The fusion block figure can be found in the supplementary material. Sample features extracted from  FCD and CCD blocks are illustrated in Fig \ref{featresponse-fcsd}.

\subsection{Complementary Nature of Shadow Removal and Detection}

\label{comp}
\begin{figure}
	\centering
	\includegraphics[width=1\linewidth]{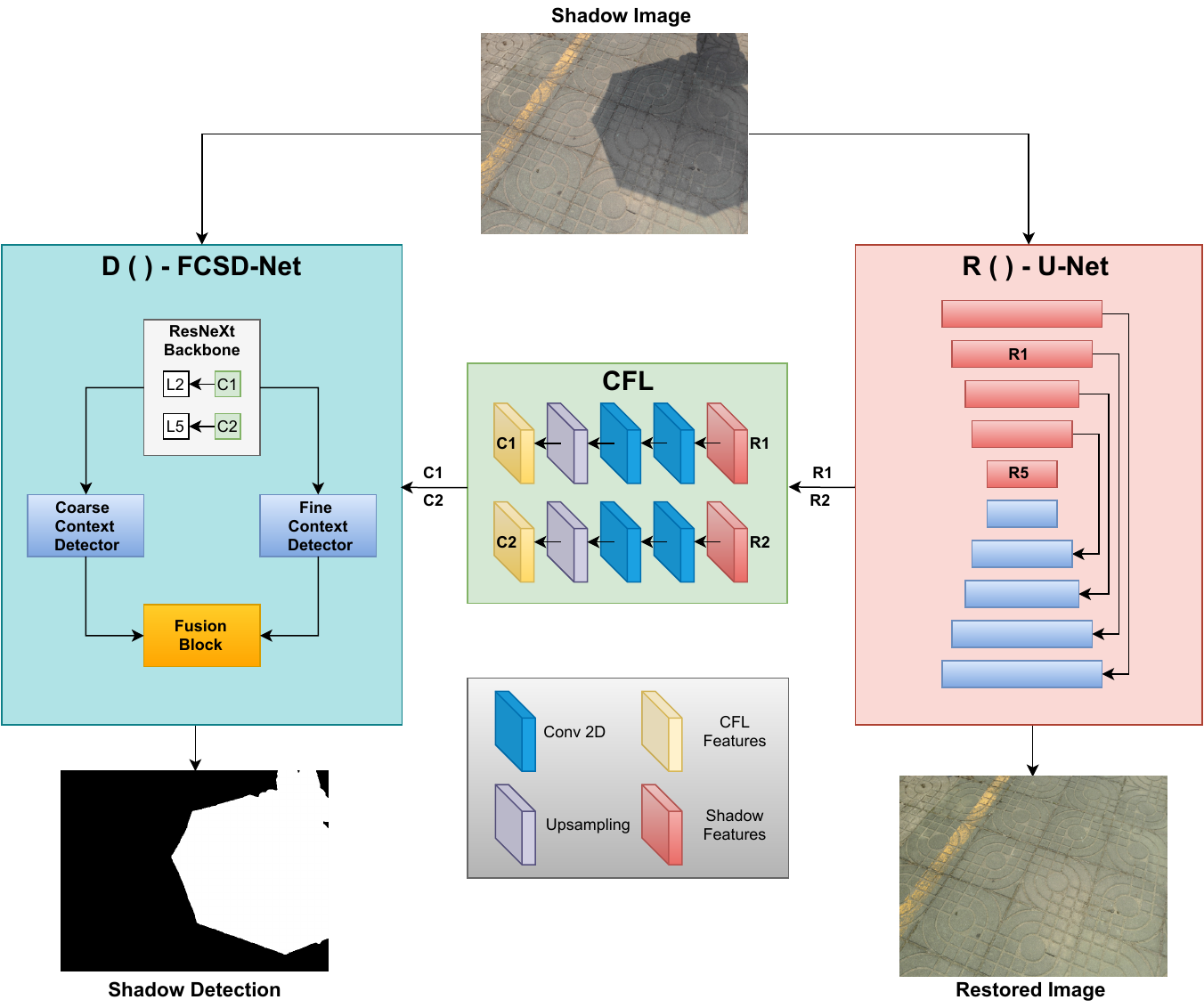}
	\caption{A detailed illustration of the proposed method- R2D. The input image is passed to both restoration network ($R ()$) - U-Net and detection network ($D ()$) - FCSD-Net. The features R1 and R2 are forwarded to CFL. The output of CFL - C1 and C2 are forwarded to FCSD-Net where they are added to L2 and L5, respectively. The output  is taken from FCSD-Net.}
	\label{R2D}
	\vspace{-0.5em}
\end{figure}

\begin{figure}[htp!]
	\centering
	\includegraphics[width=1\linewidth]{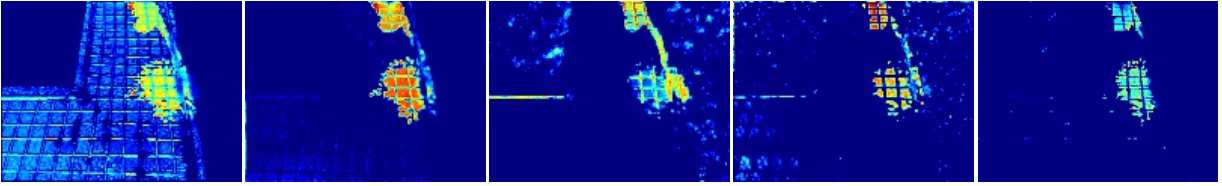}
	\caption{Feature maps collected while training U-Net for shadow removal. Even though the task here is shadow removal, the ConvNet still learns a lot of features to detect the shadow region.}
	\label{featresponse-r2d}
	\vspace{-0.5em}
\end{figure}

Shadow detection is the task of predicting the shadow binary mask from the input image with shadow. Shadow removal is the task of removing the shadow from the image and produce a restored image where the shadow part is replaced with the background. Both of these tasks need information about the shadow region as they involve careful delineation of the shadow.  To show the complementary nature of shadow removal and detection tasks, let us assume that the input image with shadow is denoted by $I$, shadow segmentation mask that has the detected shadow region is denoted $d$ and the restored image is denoted by $r$. Let the task of shadow detection be learned by a ConvNet $D ()$ and the task of shadow removal be learned by a ConvNet $R ()$. Then, the shadow detection and removal tasks can be formulated as follows: 
\setlength{\belowdisplayskip}{0pt} \setlength{\belowdisplayshortskip}{0pt}
\setlength{\abovedisplayskip}{0pt} \setlength{\abovedisplayshortskip}{0pt}
\begin{equation}
d = D(I), \;\;\;r = R(I).
\end{equation}
The shadow removal task can be further formulated as 
\begin{equation}
r = I - d, 
\end{equation}
where $d$ represents the shadow pixels that need to be removed. One can clearly see that learning $R()$ involves learning $d$.  Hence, we try to use this complementary information when we learn to restore and use it to enhance the performance of detection. This phenomenon can be empirically observed in Fig~\ref{featresponse-r2d}. 

Although the standard datasets used for shadow detection consists of only pairs of shadow images with their corresponding segmentation masks, the recently released ISTD dataset consists of triplets of images, image with shadow, shadow segmentation and the clean restored image. This opens the door for us to explore the idea of availing the information learned while learning to remove a shadow and use it for detection. 

\subsection{R2D: Restore to Detect}

In the proposed Restore to Detect (R2D) method, we have two networks $D()$ and $R()$ learning to do shadow detection and shadow removal, respectively. For our network $R()$, we use a standard U-Net \cite{ronneberger2015u} architecture as it is light-weight and effective compared to the other standard networks. U-Net consists of an encoder-decoder architecture where both the encoder and decoder have 5 conv blocks each. There are skip connections between the  conv blocks in the encoder to the conv blocks of the decoder. We select the output feature maps from the 2nd layer and the 5th layer of the encoder to pass them to network $D()$. Note that though all the feature maps at each level can be forwarded to the network $D()$, we choose to forward the feature maps at only 2 levels as from our experiments we found that it was enough in terms of performance and also helped to reduce the complexity of our method. We choose the 2nd layer and the 5th layer specifically as  we could leverage both local information and global information learned by the network $R()$, respectively. We term these feature maps $R1$ and $R2$. More explanation on why we exactly choose the 2nd and the 5th layer can be found in the supplementary material. Note that we did not use any state-of-the-art shadow removal network in $R()$ as it increases the complexity by a lot and hinders while optimizing $D()$ in the R2D framework.

We propose a Complementary Feature Learning (CFL) block to efficiently learn and forward only the features that correspond to shadow regions. In CFL, we have conv layers acting at each feature map level, $R1$ and $R2$. We then interpolate them to match the size of the feature maps where it is going to be added in the $D()$ network. We term these features $C1$ and $C2$. We fuse $C1$ and $C2$ with $L2$ and $L5$, respectively in FCSD-Net. The R2D learning strategy is illustrated in Fig \ref{R2D}. 

\subsection{Training Strategy}

\begin{table*}[]
	\centering
	\resizebox{2\columnwidth}{!}{
		\begin{tabular}{
				>{\columncolor[HTML]{FFFFFF}}c |
				>{\columncolor[HTML]{FFFFFF}}c 
				>{\columncolor[HTML]{FFFFFF}}c 
				>{\columncolor[HTML]{FFFFFF}}c |
				>{\columncolor[HTML]{FFFFFF}}c 
				>{\columncolor[HTML]{FFFFFF}}c 
				>{\columncolor[HTML]{FFFFFF}}c |
				>{\columncolor[HTML]{FFFFFF}}c 
				>{\columncolor[HTML]{FFFFFF}}c 
				>{\columncolor[HTML]{FFFFFF}}c |
				>{\columncolor[HTML]{FFFFFF}}c }
			{\color[HTML]{000000} Method} & \multicolumn{3}{c|}{\cellcolor[HTML]{FFFFFF}{\color[HTML]{000000} UCF \cite{zhu2010learning}}} & \multicolumn{3}{c|}{\cellcolor[HTML]{FFFFFF}{\color[HTML]{000000} SBU \cite{vicente2016large}}} & \multicolumn{3}{c|}{\cellcolor[HTML]{FFFFFF}{\color[HTML]{000000} ISTD \cite{wang2018stacked}}} & Mean \\ \hline
			& BER & Shadow & Non shad. & BER & Shadow & Non shad. & BER & Shadow & Non shad. & BER \\ \hline
			$MTMT$-Net (CVPR 2020) \cite{chen2020multi} & 7.47 & 10.31 & 4.63 & 3.15 & 3.73 & 2.57 & 1.72 & 1.36 & 2.08 & 4.11 \\ 
			{\color[HTML]{000000} $stacked$-$CNN$ (ECCV 16) \cite{vicente2016large}} & {\color[HTML]{000000} 13.00} & {\color[HTML]{000000} 9.00} & {\color[HTML]{000000} 17.1} & {\color[HTML]{000000} 10.8} & {\color[HTML]{000000} 8.84} & {\color[HTML]{000000} 12.76} & {\color[HTML]{000000} 8.6} & {\color[HTML]{000000} 7.69} & {\color[HTML]{000000} 9.23} & 10.8 \\
			{\color[HTML]{000000} $SRM$ (ICCV 17) \cite{wang2017stagewise}} & {\color[HTML]{000000} 12.51} & {\color[HTML]{000000} 21.41} & {\color[HTML]{000000} 3.6} & {\color[HTML]{000000} 6.51} & {\color[HTML]{000000} 10.52} & {\color[HTML]{000000} 2.50} & {\color[HTML]{000000} 7.92} & {\color[HTML]{000000} 13.97} & {\color[HTML]{000000} 1.86} & 8.98 \\
			{\color[HTML]{000000} $scGAN$ (ICCV 17) \cite{nguyen2017shadow}} & {\color[HTML]{000000} 11.52} & {\color[HTML]{000000} 7.74} & {\color[HTML]{000000} 15.3} & {\color[HTML]{000000} 9.04} & {\color[HTML]{000000} 8.39} & {\color[HTML]{000000} 9.69} & {\color[HTML]{000000} 4.7} & {\color[HTML]{000000} 3.22} & {\color[HTML]{000000} 6.18} & 8.42 \\
			{\color[HTML]{000000} $ST$-$CGAN$ (CVPR 18) \cite{wang2018stacked}} & {\color[HTML]{000000} 11.23} & {\color[HTML]{000000} 4.94} & {\color[HTML]{000000} 17.52} & {\color[HTML]{000000} 8.14} & {\color[HTML]{000000} 3.75} & {\color[HTML]{000000} 12.53} & {\color[HTML]{000000} 3.85} & {\color[HTML]{000000} 2.14} & {\color[HTML]{000000} 5.55} & 7.74 \\
			{\color[HTML]{000000} $DSC$ (CVPR 18) \cite{hu2018direction}} & {\color[HTML]{000000} 10.54} & {\color[HTML]{000000} 18.08} & {\color[HTML]{000000} 3.00} & {\color[HTML]{000000} 5.59} & {\color[HTML]{000000} 9.76} & {\color[HTML]{000000} 1.42} & {\color[HTML]{000000} 3.42} & {\color[HTML]{000000} 3.85} & {\color[HTML]{000000} 3.00} & 6.51 \\
			{\color[HTML]{000000} $ADNet$ (ECCV 18) \cite{le2018a+}} & {\color[HTML]{000000} 9.25} & {\color[HTML]{000000} 8.37} & {\color[HTML]{000000} 10.14} & {\color[HTML]{000000} 5.37} & {\color[HTML]{000000} 4.45} & {\color[HTML]{000000} 6.3} & {\color[HTML]{000000} -} & {\color[HTML]{000000} -} & {\color[HTML]{000000} -} & 7.31 \dag \\
			{\color[HTML]{000000} $RAS$ (ECCV 18) \cite{chen2018reverse}} & {\color[HTML]{000000} 13.62} & {\color[HTML]{000000} 23.06} & {\color[HTML]{000000} 4.18} & {\color[HTML]{000000} 7.31  } & {\color[HTML]{000000} 12.13} & {\color[HTML]{000000} 2.48} & {\color[HTML]{000000} 11.14} & {\color[HTML]{000000} 19.88} & {\color[HTML]{000000} 2.41} & 10.69 \\
			$BDRAR$ (ECCV 18) \cite{zhu2018bidirectional} & 7.81 & 9.69 & 5.94 & 3.64 & 3.40 & 3.89 & 2.69 & 0.50 & 4.87 & 4.71 \\
			{\color[HTML]{000000} $DSDNet$ (CVPR 19) \cite{zheng2019distraction}} & {\color[HTML]{000000} 7.59} & {\color[HTML]{000000} 9.74} & {\color[HTML]{000000} 5.44} & {\color[HTML]{000000} 3.45} & {\color[HTML]{000000} 3.33} & {\color[HTML]{000000} 3.58} & {\color[HTML]{000000} 2.17} & {\color[HTML]{000000} 1.36} & {\color[HTML]{000000} 2.98} & 4.40 \\
			$DC$-$DSPF$ (IJCAI 19) \cite{wang2019densely} & 7.90 & 6.50 & 9.30 & 4.90 & 4.70 & 5.10 & - & - & - & 6.40 \dag \\
			$EGNet$ (ICCV 19) \cite{zhao2019egnet}& 9.20 & 11.28 & 7.12 & 4.49 & 5.23 & 3.75 & 1.85 & 1.75 & 1.95 & 5.11 \\ \hline
			
			{\color[HTML]{000000} $R2D$ (Ours)} & {\color[HTML]{000000} \textbf{6.96}} & {\color[HTML]{000000} \textbf{8.32}} & {\color[HTML]{000000} \textbf{5.60}} & {\color[HTML]{000000} \textbf{3.15}} & {\color[HTML]{000000} \textbf{2.74}} & {\color[HTML]{000000} \textbf{3.56}} & {\color[HTML]{000000} \textbf{1.69}} & {\color[HTML]{000000} \textbf{0.59}} & {\color[HTML]{000000} \textbf{2.79}} & \textbf{3.93}
		\end{tabular}
	}
	\vskip -10 pt
	\caption{Comparison of quantitative results in terms of BER of shadow region, non-shadow region and average BER for the UCF, SBU and ISTD datasets with the state-of-the-art shadow detectors. Note that all the numbers are in terms of error, so the lesser the better. \dag Mean found only using 2 datasets. Our results are in \textbf{Bold} and we acheive state-of-the art across all datasets with least BER.  }
	\label{quanres}
	\vspace{-0.5em}
\end{table*}

For training R2D, we jointly optimize the shadow prediction, false positive (FP) and false negative (FN) maps at all scales. Note that all the datasets provide ground truth shadow segmentation masks. For FP and FN ground truth maps, we use the data provided by \cite{zheng2019distraction} which was created based on the differences between the existing shadow detection and their ground truths. We first find a weighted binary cross entropy loss $\mathcal{L}_W$ as follows:
\begin{equation*}
\mathcal{L}_W = -\left( \sum_{i} (a_{n} y_i \log(x_i) + b_{n} (1-y_i) \log(1-x_i) )\right),	
\end{equation*}
where $x_i$ corresponds to the prediction, $y_i$ corresponds to the ground-truth, $a_{n}=\frac{N_n}{N_n + N_p}$, $b_{n}=\frac{N_p}{N_n + N_p}$. $N_n$ and $N_p$ are the number of positive pixels and negative pixels, respectively. Note that this weighted loss is applied pixel wise and summed up over all the pixels. We calculate $\mathcal{L}_W$ for FP and FN maps and term them $\mathcal{L}_{FP}$ and $\mathcal{L}_{FN}$, respectively. We also adopt the distraction-aware cross entropy loss $\mathcal{L}_{DS}$ from \cite{zheng2019distraction} which forces the predictions to be less prone to false detections. $\mathcal{L}_{DS}$ is formulated as follows:
\begin{equation}
\begin{multlined}
\mathcal{L}_{DS} = -( \sum_{i} (a_{n} y_i^{fnd} y_i \log(x_i) \\
+ b_{n} y_i^{fpd} (1-y_i) \log(1-x_i) ) ),	
\end{multlined}
\end{equation}
where $y_i^{fpd}$ is the ground truth of FP pixel and $y_i^{fnd}$ is the ground truth of FN pixel. We define the shadow loss $\mathcal{L}_{shadow}$ calculated on the final shadow prediction as:  
\begin{equation}
\mathcal{L}_{shadow} = \mathcal{L}_{W} +     \mathcal{L}_{W}\mathcal{L}_{DS},
\end{equation} 
which is calculated between the shadow predictions and the ground truth segmentation map. The total loss used to train the detection network is defined as follows:
\begin{equation}
\mathcal{L}_{det} =  \sum_k \alpha \mathcal{L}_{shadow}^k +    \beta \mathcal{L}_{FP}^k +  \gamma \mathcal{L}_{FN}^k,
\end{equation} 
where $k$ represents the shadow predictions found across different scales in the fusion block. The values of $\alpha, \beta$ and $\gamma$ are set equal to 1, 2 and 2, respectively in our experiments.  The $L_{2}$ loss is used to pretrain the restoration network (U-Net).  It is defined as: $\mathcal{L}_{res} = \sum_i (y_i - p_i)^2,$
where $y_i$ corresponds to the target clean image pixel and $p_i$ corresponds to the prediction at the $i^{th}$ pixel location in the restored image.  During the fine-tuning stage, we train the entire framework using the following total loss:
\begin{equation}
\mathcal{L}_{R2D} = \mathcal{L}_{det} + \mathcal{L}_{res}.
\end{equation}

While using the R2D strategy, we first pretrain the restoration network $R()$ on the ISTD dataset using $\mathcal{L}_{res}$ for 500 epochs. $R()$ performs restoration reasonably well as it achieves an RMSE of 11.21 while evaluated on ISTD test dataset. For training the images we first rescale them to $320 \times 320$ resolution. We use a batch size of 16 and a learning rate of 0.001. We use SGD as the optimizer with momentum set at 0.9 and weight decay at 0.0001. We use the PyTorch framework for training using a Nvidia Quadro RTX-8000 GPU. Note that the ResNext-101 was pre-trained on ImageNet and all the other parameters are randomly initialized. We also use data augmentation by randomly flipping both the input and the ground truth.

Note that only the ISTD dataset consists of triplets of images while the SBU and UCF datasets contain only image pairs. The pretraining stage is common for experiments on all the datasets where the restoration network is trained on the ISTD dataset. During the finetuning phase while training ISTD dataset, we still have access to the clean images so we train the entire network with   $\mathcal{L}_{R2D}$. For the other datasets, during the fine tuning stage we only use $\mathcal{L}_{det}$. Note that even then, we have feature maps flowing from the restoration network to the detection network which enhance the quality of shadow predictions. The weights are also back-propagated through the CFL to the encoder of the restoration network which makes our training strategy not dependent on the availability of clean images after pretraining. The fine-tuning is performed for 6000 iterations.

\subsection{Inference Strategy}
During inference , we feed forward the image to both FCSD-Net and U-Net ($D()$ and $R()$ respectively). The output shadow segmentation map is taken from $D()$. We do stochastic weight averaging \cite{izmailov2018averaging} on the models saved at 4000, 5000, and 6000 iterations and use it to perform the inference. 

\begin{figure*}
	\centering
	\includegraphics[width=1\linewidth]{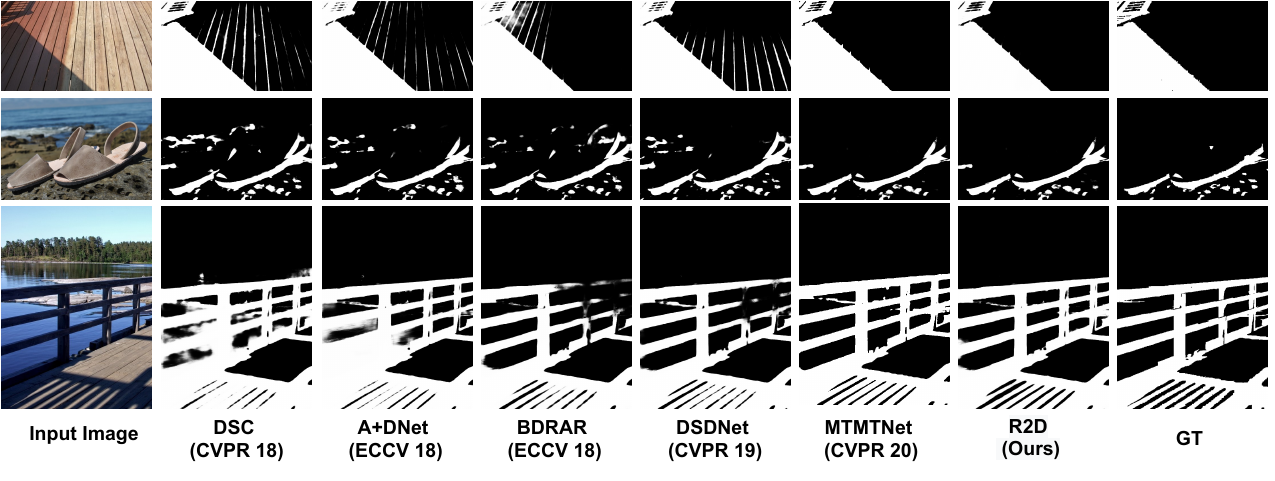}
	\vskip-12pt\caption{Comparison of predictions of our proposed methods with leading shadow detection methods. The first and last columns correspond to the input and ground-truth, respectively. Other columns correspond to the predictions obtained using different methods. }
	\label{qualres}
	\vspace{-0.5em}
\end{figure*}
\section{Experimental and Results}
In this section, we give details about the experiments we conduct to compare our method against the recent deep learning-based methods: stacked-CNN \cite{vicente2017leave}, SRM \cite{wang2017stagewise}, scGAN \cite{nguyen2017shadow}, ST-GAN \cite{wang2018stacked}, BDRAR \cite{zhu2018bidirectional}, DSC \cite{hu2018direction}, ADNet \cite{le2018a+}, DC-DSPF \cite{wang2019densely},  RAS \cite{chen2018reverse}, MTMT-Net \cite{chen2020multi} and DSDNet \cite{zheng2019distraction}. We illustrate the qualitative results as well as compute the performance metrics that are widely used in the shadow detection literature to quantitatively compare our proposed methods with the recent methods. 


\subsection{Datasets and Evaluation Metrics}
The following three datasets are used to conduct experiments -
UCF \cite{zhu2010learning}, SBU \cite{vicente2016large} and ISTD \cite{wang2018stacked} datasets. In both the SBU and UCF datasets, there are images with shadows and corresponding shadow segmentation masks. The SBU dataset consists of 4089 training images and 638 testing images. In the UCF dataset, we only use the testing set which contains 110 images similar to the previous works. We train the network on the SBU training set and test on both SBU testing set and UCF testing set. Unlike the SBU and UCF datasets, ISTD dataset contains triplets of images- image with shadow, shadow segmentation mask and clean image without shadow. The number of training and testing sets of images in ISTD is 1870 and 540 respectively.

We use the balanced error rate (BER) as the performance metric for quantitative comparisons with the recent methods. BER is calculated as follows:
\begin{equation}
BER = 1 - \frac{1}{2}  (\frac{TP}{TP+FN} + \frac{TN}{TN+FP}),
\end{equation}
where TP, TN, FP and FN correspond to the number of pixels which are true positives, true negatives, false positives and false negatives, respectively. We calculate the BER of shadow and the non-shadow regions separately and then calculate the average BER. We report the individual shadow, non-shadow BER as well as the average BER. We also report the mean BER found across all 3 datasets.

\begin{figure*}
	\centering
	\includegraphics[width=1\linewidth]{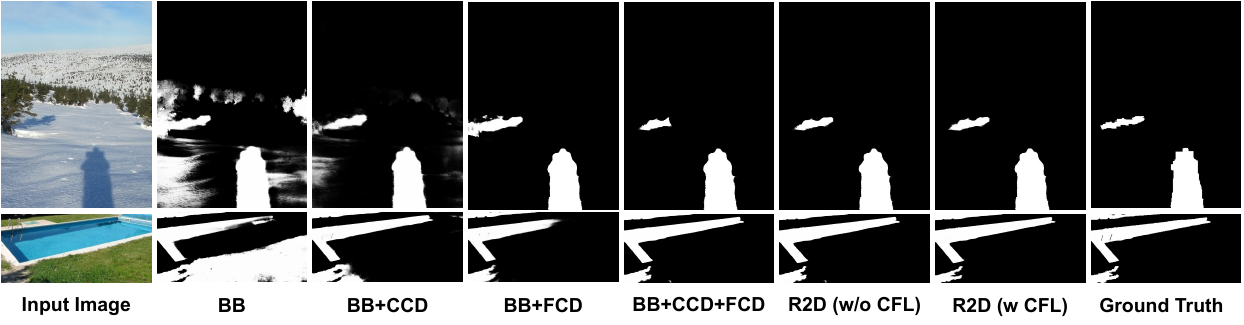}
	\vskip-8pt	\caption{Comparison of qualitative results for ablation study. The first and last column correspond to the input and ground truth respectively. Other columns correspond to the predictions obtained using different configurations in the ablation study. 
	\vspace{-2em}
	}
	\label{ablres}
\end{figure*}

\subsection{Quantitative Results}
In Table \ref{quanres}, we summarize the results of our experiments. It can be observed from this table that our methods are observed to perform the best across all 3 datasets. We note that overall MTMT-Net \cite{chen2020multi} and DSDNet \cite{zheng2019distraction} are the second best performing methods.  When compared to DSDNet, our method has an 8.30 \%, 8.69 \%, 22.11 \% improvement in terms of BER on the UCF, SBU and ISTD datasets, respectively. We also calculate the mean BER across all datasets where we achieve an improvement of 10.68 \%, 4.3\% over DSDNet and MTMT-Net, respectively. In addition, we also conducted an experiment using the residuals as the shadow segmentation masks as explained in Section~\ref{intro}. The performance was not even comparable with any of the baseline methods as the BER on the ISTD dataset was 27.32.  

\subsection{Qualitative Results}
We visualize the predictions of our method and the best performing methods for comparison in Fig \ref{qualres}.  It can be observed that our method's predictions are better and more closer to the ground truth when compared to other methods. In the first row, other methods mistake the lines between the wooden strips as shadows as they are of the same color of the shadows. However, our method does not falsely detect them as shadows as we focus on extracting fine context shadow regions better. Similar observations can be made in the third row. In the second row, our method does not falsely detect the smaller shadow regions found in the shore while all the other methods detect them as shadows. 

\section{Discussion}

\noindent \textbf{Ablation Study:} We conduct an ablation study to show the importance of each individual component in our proposed method. We start with just using the backbone features (BB) for shadow detection. Then, we add the CCD block and the FCD block separately and conduct experiments. Then we use both blocks together and use the fusion block to fuse the features learned by CCD and FCD. This configuration corresponds to FCSD-Net. We  then add the U-Net based restoration network  and directly fuse the feature maps to FCSD-Net. Then, we add the CFL block to specifically learn shadow features to be forwarded from the restoration network to the detection network. This configuration corresponds to  R2D. The results corresponding to these experiments are shown in Table \ref{abl} and Figure \ref{ablres}.

\noindent \textbf{Using R2D with other networks:} Also, to show the adaptability of R2D, we conduct an experiment where we  use R2D for DSDNet. The results can be seen in Table \ref{ablr2d}. It can be observed that we get an improvement of 5.90 \% while using R2D learning strategy to train DSDNet when compared to training it normally. 

\begin{figure}
	\centering
	\includegraphics[width=1\linewidth]{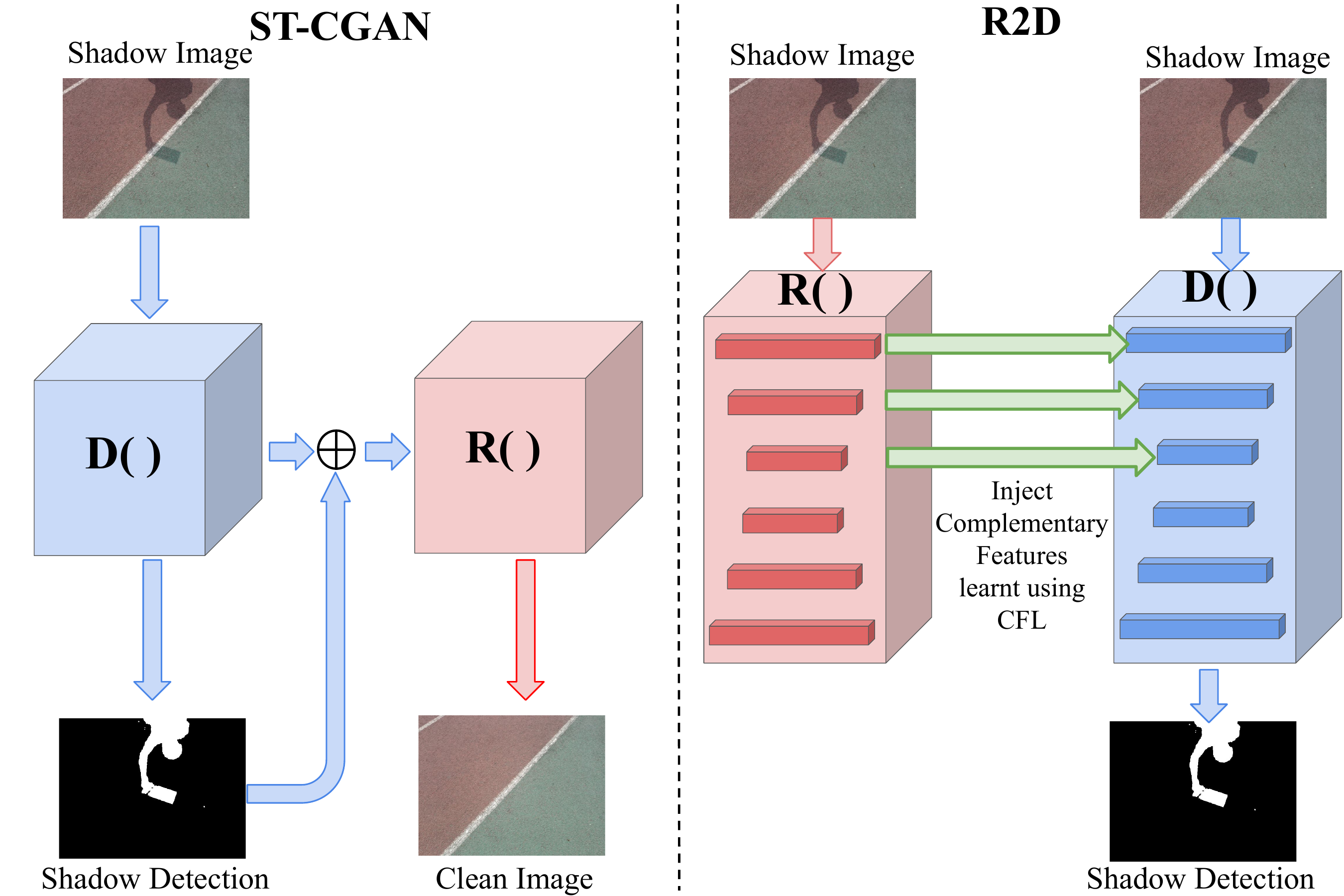}
	\caption{The main differences between ST-cGAN and R2D.}
	\label{diff}
	\vspace{-0.5em}
\end{figure}

\noindent \textbf{Difference from ST-cGAN:} ST-cGAN \cite{wang2018stacked} uses a stacked conditional generative adversarial network to solve shadow detection and removal in sequence. It uses the shadow mask output as an additional information for the shadow removal network. Note that the detection pipeline in ST-cGAN is generic. In contrast, R2D learning strategy uses complementary feature maps information from the removal network to improve the detection performance. This difference is also visualized in Figure \ref{diff}.

\begin{table}[]
	\centering
	\resizebox{1\columnwidth}{!}{
		\begin{tabular}{
				>{\columncolor[HTML]{FFFFFF}}c |
				>{\columncolor[HTML]{FFFFFF}}c |
				>{\columncolor[HTML]{FFFFFF}}l |
				>{\columncolor[HTML]{FFFFFF}}l |
				>{\columncolor[HTML]{FFFFFF}}c |
				>{\columncolor[HTML]{FFFFFF}}l |
				>{\columncolor[HTML]{FFFFFF}}l |
				>{\columncolor[HTML]{FFFFFF}}c 
				>{\columncolor[HTML]{FFFFFF}}l 
				>{\columncolor[HTML]{FFFFFF}}l |
				>{\columncolor[HTML]{FFFFFF}}c }
			{\color[HTML]{000000} Method} & \multicolumn{3}{c|}{\cellcolor[HTML]{FFFFFF}{\color[HTML]{000000} UCF \cite{zhu2010learning}}} & \multicolumn{3}{c|}{\cellcolor[HTML]{FFFFFF}{\color[HTML]{000000} SBU \cite{vicente2016large}}} & \multicolumn{3}{c|}{\cellcolor[HTML]{FFFFFF}{\color[HTML]{000000} ISTD \cite{wang2018stacked}}} & Mean \\ \hline
			BB & \multicolumn{3}{c|}{\cellcolor[HTML]{FFFFFF}9.64} & \multicolumn{3}{c|}{\cellcolor[HTML]{FFFFFF}4.48} & \multicolumn{3}{c|}{\cellcolor[HTML]{FFFFFF}3.41} & 5.84 \\
			BB + CCD & \multicolumn{3}{c|}{\cellcolor[HTML]{FFFFFF}7.63} & \multicolumn{3}{c|}{\cellcolor[HTML]{FFFFFF}3.48} & \multicolumn{3}{c|}{\cellcolor[HTML]{FFFFFF}2.18} & 4.43 \\
			{\color[HTML]{000000} BB + FCD} & \multicolumn{3}{c|}{\cellcolor[HTML]{FFFFFF}{\color[HTML]{000000} 7.52}} & \multicolumn{3}{c|}{\cellcolor[HTML]{FFFFFF}{\color[HTML]{000000} 3.40}} & \multicolumn{3}{c|}{\cellcolor[HTML]{FFFFFF}{\color[HTML]{000000} 1.82}} & 4.24 \\
			{\color[HTML]{000000} BB +CCD + FCD } & \multicolumn{3}{c|}{\cellcolor[HTML]{FFFFFF}{\color[HTML]{000000} 7.08}} & \multicolumn{3}{c|}{\cellcolor[HTML]{FFFFFF}{\color[HTML]{000000} 3.30}} & \multicolumn{3}{c|}{\cellcolor[HTML]{FFFFFF}{\color[HTML]{000000} 1.71}} & 4.03 \\
			{\color[HTML]{000000}  R2D (w/o CFL)} & \multicolumn{3}{c|}{\cellcolor[HTML]{FFFFFF}{\color[HTML]{000000} 7.04}} & \multicolumn{3}{c|}{\cellcolor[HTML]{FFFFFF}{\color[HTML]{000000} 3.26}} & \multicolumn{3}{c|}{\cellcolor[HTML]{FFFFFF}{\color[HTML]{000000} 1.70}} & 4.00 \\
			{\color[HTML]{000000}  R2D (w CFL)} & \multicolumn{3}{c|}{\cellcolor[HTML]{FFFFFF}{\color[HTML]{000000} 6.96}} & \multicolumn{3}{c|}{\cellcolor[HTML]{FFFFFF}{\color[HTML]{000000} 3.15}} & \multicolumn{3}{c|}{\cellcolor[HTML]{FFFFFF}{\color[HTML]{000000} 1.69}} & 3.93
		\end{tabular}
	}
	\vskip -10 pt
	\caption{Comparison of quantitative results in terms of average BER for ablation study. }
	\label{abl}
	\vspace{-0.5em}
\end{table}

\begin{table}[]
	\centering
	\resizebox{0.9\columnwidth}{!}{
		\begin{tabular}{
				>{\columncolor[HTML]{FFFFFF}}l |
				>{\columncolor[HTML]{FFFFFF}}l |
				>{\columncolor[HTML]{FFFFFF}}l |
				>{\columncolor[HTML]{FFFFFF}}l |
				>{\columncolor[HTML]{FFFFFF}}l }
			
			Method & UCF \cite{zhu2010learning} & SBU \cite{vicente2016large}& ISTD \cite{wang2018stacked} & Mean \\ \hline
			DSDNet & 7.59 & 3.45 & 2.17 & 4.40 \\
			DSDNet + R2D & 7.30 & 3.39 & 1.78 & 4.15
		\end{tabular}
	}
	\vskip -10 pt
	\caption{Comparison of quantitative results in terms of average BER for ablation study on using R2D for DSD-Net.}
	\label{ablr2d}
	\vspace{-0.5em}
\end{table}

\noindent \textbf{Fine Context Feature Learning:} A generic ConvNet has a encoder-decoder architecture which is an undercomplete type of architecture spatially that learns more high level features when the network is designed more deep.  Overcomplete representations \cite{lewicki2000learning} were initially introduced in signal processing as an alternate method for signal representation. Overcomplete bases or dictionaries were proposed where the number of basis functions are more than the number of samples of input signal. Overcomplete bases have a better flexibility at capturing the structure of the data and so is more robust. In \cite{vincent2008extracting}, overcomplete auto-encoders were observed to be better feature extractors for denoising. In an overcomplete auto-encoder, the number of neurons in the hidden layer is more than the that of the initial layers. So typically, the dimensionality of the representation in the deeper layers is more than that of the input layer. In the deep learning era, the concept of overcomplete representations has been under-explored \cite{valanarasu2020kiu, yasarla2020exploring, valanarasu2020kiu1}. In an overcomplete alternate convolutional network the input image is taken to a higher dimension spatially. So, the max-pooling layers in a typical ConvNet can be replaced with upsampling layers to prevent the receptive field size to increase in the deeper layers of the network. 

Consider a configuration of two conv layers in succession where $I$ be the input image, $F_1$ and $F_2$ be the feature maps extracted from the conv layers 1 and 2, respectively. Let the initial receptive field of the conv filter be $k \times k$ on the image. Now, if there is a max-pooling layer present in between the conv layers like in generic ConvNets, the receptive field would become larger in the successive layers. The receptive field size change due to max-pooling layer is dependent on two variables- pooling coefficient and stride of the pooling filter. Considering a default configuration (like in most cases) where both pooling coefficient and stride is 2, the receptive field of conv layer 2 (to which $F_1$ is forwarded) on the input image would be $ 2 \times k \times 2 \times k$. Similarly, the receptive field of conv layer 3 (to which $F_2$ is forwarded) would be $ 4 \times k \times 4 \times k$. This increase in receptive field can be generalized for the $i^{th}$ layer in an undercomplete network as follows:

\begin{equation}
RF (w.r.t \; I) =  2^{2*(i-1)} \times k \times k
\end{equation}

In an overcomplete ConvNet, we propose using an upsampling layer instead of the max-pooling layer. As the upsampling layer actually works opposite to that of max-pooling layer, the receptive field of conv layer 2 on the input image now would be $ \frac{1}{2} \times k \times \frac{1}{2} \times k$. Similarly, the receptive field of conv layer 3  now would be $ \frac{1}{4} \times k \times \frac{1}{4} \times k$. This increase in receptive field can be generalized for the $i^{th}$ layer in the overcomplete ConvNet as follows:  

\begin{equation}
RF (w.r.t \; I) =  \left(\frac{1}{2}\right)^{2*(i-1)} \times k \times k. \end{equation}

This helps in an overcomplete network to learn more low-level information like edges and other finer details better. So in our work, the Fine Context Block has this alternate ConvNet architecture to learn fine details of the shadow region. 

\noindent \textbf{FCD architecture Design Justification:} In the Fine Context Detector (FCD) block, we  use four convolutional blocks where we upsample the input features at each block. Each conv block has a conv layer followed by an upsampling layer and ReLU activation. The upsampling layer used here is bilinear interpolation. We use four conv blocks such that the final resolution of the feature map at FCD block is $400 \times 400$. This limit is based on trade-off between performance and model complexity. We noticed that the change in performance was not too affected after we reached a resolution of $400 \times 400$. This observation could be explained as with increase in resolution the complexity of the network training gets hindered. We conducted experiments on ISTD dataset to show how the number of conv blocks in FCD block affected the performance. These observations can be found in Table \ref{tab:my-table}. 

\begin{table}[]
	\centering
	\begin{tabular}{
			>{\columncolor[HTML]{FFFFFF}}c |
			>{\columncolor[HTML]{FFFFFF}}c |
			>{\columncolor[HTML]{FFFFFF}}c |
			>{\columncolor[HTML]{FFFFFF}}c |
			>{\columncolor[HTML]{FFFFFF}}c |
			>{\columncolor[HTML]{FFFFFF}}c }
		
		No. of conv blocks & 1 & 2 & 3 & 4 & 5 \\ \hline
		BER & 2.10 & 1.85 & 1.77 & \textbf{1.71} & \textbf{1.71}
	\end{tabular}
	\caption{Change in performance with the number of conv blocks in the FCD block. The performance saturates after 4 number of conv blocks.}
	\label{tab:my-table}
\end{table}

\noindent \textbf{Choosing R1 and R2:} In our proposed framework, we chose $R1$ and $R2$ from the second and last layer of encoder in U-Net respectively. We do not feed forward all the features from $R ()$ to $D ()$ as it increases the complexity of network. The motivation is to feed forward a good combination of both local and global features of the shadow region to the detection network $D ()$. We observe that features at layer 1 and 2 extract local features while layers 3,4, and 5 extract global features. So we choose $R1$ and $R2$ from layers 2 and 5 to get the abstract local and global features from network $R ()$. 

\noindent \textbf{Limitations:} We acknowledge that our method is more computationally complex (1.3 $\times$ DSDNet in terms of FLOPs) than previous methods. Our inference time is 0.26 seconds per image of resolution $320 \times 320$ which is more than MTMT-Net. In the R2D framework, we used  U-Net as our restoration network. A better strategy would be to use powerful shadow removal networks like DHAN \cite{cun2020towards}. From our experiments, we were unable to train $R()$ and $D()$ efficiently when we chose $R()$ as DHAN instead of U-Net due to the huge number of parameters and difficulty in optimizing both networks in parallel. With better engineering strategy, using DHAN should be able to further improve the effectivess of R2D. Using FCSD-Net's shadow detection features to boost shadow removal performance is also a possible direction left unexplored in this work due to lack of compute power. Optimizing the networks in parallel and leveraging shadow removal for shadow detection and vice-versa is another possible setup not explored in this paper. Although these setups are theoretically possible, the road-blocks are compute power and unstable training. 
\section{Conclusion}
In this work, we explored a new direction for shadow detection. We propose a new method, R2D, in which we leverage the shadow features learned during shadow removal to improve the shadow detection performance. We also propose a new network architecture, FCSD-Net, that serves as the detection network architecture in our R2D framework. It mainly focuses on fine context feature extraction for shadow detection. We do this by designing a fine context detector block where we constrain the receptive field size to focus more on the local features which improves the detection performance especially in confounding cases where the shadow region and background have similar color intensities. R2D can be easily adopted as the learning strategy to enhance any shadow detection network. We conduct extensive experiments to show the effectiveness of our methods. Using shadow detection as an auxiliary task for shadow removal is considered as a future direction of this work.

{\small
\bibliographystyle{ieee_fullname}
\bibliography{egbib}
}

\end{document}